\documentclass[sigconf]{acmart}
\usepackage{multirow}
\usepackage{lineno}
\pdfoutput=1
\AtBeginDocument{%
  \providecommand\BibTeX{{%
    \normalfont B\kern-0.5em{\scshape i\kern-0.25em b}\kern-0.8em\TeX}}}

\copyrightyear{2022}
\acmYear{2022}
\setcopyright{acmcopyright}\acmConference[ICVGIP'22]{Proceedings of the Thirteenth Indian Conference on Computer Vision, Graphics and Image Processing}{December 8--10, 2022}{Gandhinagar, India}
\acmBooktitle{Proceedings of the Thirteenth Indian Conference on Computer Vision, Graphics and Image Processing (ICVGIP'22), December 8--10, 2022, Gandhinagar, India}
\acmPrice{15.00}
\acmDOI{10.1145/3571600.3571608}
\acmISBN{978-1-4503-9822-0/22/12}

\date{}
\usepackage{nopageno}
\title{Masked Student Dataset of Expressions}
\hypersetup{draft}
\begin{document}

\author{Sridhar Sola}
\affiliation{%
  \institution{Sri Sathya Sai Institute of Higher Learning}
  \streetaddress{}
  \city{Bengaluru}
  \state{Karnataka}
  \country{India}}
\email{solasridhar@gmail.com}

\author{Darshan Gera}
\affiliation{%
  \institution{Sri Sathya Sai Institute of Higher Learning}
  \streetaddress{}
  \city{Bengaluru}
  \state{Karnataka}
  \country{India}}
\email{darshangera@sssihl.edu.in}



\renewcommand{\shortauthors}{Sola and Gera}

\begin{abstract}
Facial expression recognition (FER) algorithms work well in constrained environments with little or no occlusion of the face. However, real-world face occlusion is prevalent, most notably with the need to use a face mask in the current Covid-19 scenario. While there are works on the problem of occlusion in FER, little has been done before on the particular face mask scenario. Moreover, the few works in this area largely use synthetically created masked FER datasets. Motivated by these challenges posed by the pandemic to FER, we present a novel dataset, the \textbf{Masked Student Dataset of Expressions} or \textbf{MSD-E}, consisting of 1,960 real-world non-masked and masked facial expression images collected from 142 individuals. Along with the issue of obfuscated facial features, we illustrate how other subtler issues in masked FER are represented in our dataset. We then provide baseline results using ResNet-18, finding that its performance dips in the non-masked case when trained for FER in the presence of masks. To tackle this, we test two training paradigms: contrastive learning and knowledge distillation, and find that they increase the model's performance in the masked scenario while maintaining its non-masked performance. We further visualise our results using t-SNE plots and Grad-CAM, demonstrating that these paradigms capitalise on the limited features available in the masked scenario. Finally, we benchmark SOTA methods on MSD-E. The dataset is available at https://github.com/SridharSola/MSD-E.
\end{abstract}

\ccsdesc[500]{Human-centered computing~HCI design and evaluation methods}

\keywords{Human-computer interaction, Computer vision, Facial expression recognition, Covid-19}

\maketitle
\nolinenumbers
\section{Introduction}
Facial expressions give hints about others' thoughts, emotions, and reactions. We humans modify our actions and words based on these facial signals. Evidently, facial expressions are a cornerstone of human-human interaction. It is not surprising then that enabling computers to recognize different facial expressions will be crucial in developing seamless human-computer interaction. Automatic Facial Expression Recognition (FER) aims to do just that.

By using Deep Learning (DL) and Convolutional Neural Networks, FER has reached a state where models are ready to be deployed in the real world \cite{article1}. However, despite the advantages of DL, FER has been plagued by issues like occlusion, pose variation, and noisy annotations. The amount of occlusion, in particular, has been exacerbated by the Covid-19 pandemic. The necessity to wear a face mask means that FER systems lose around half of facial features as masks cover the face and nose regions. Crucially, these areas have critical features needed to discriminate between emotions \cite{KOTSIA20081052}.
\begin{figure}
  \centering
  \includegraphics[scale = 0.5]{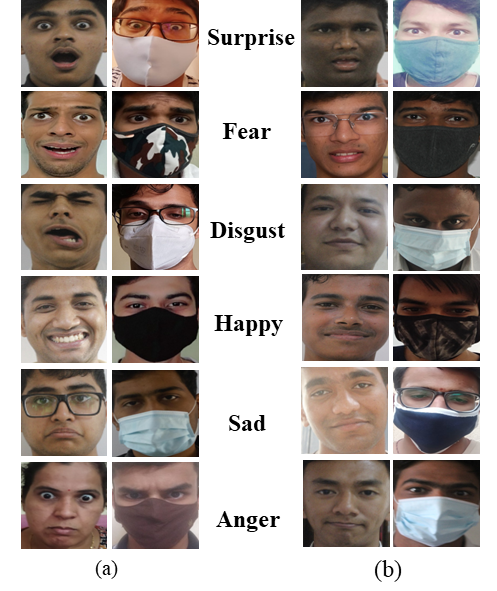}
  \caption{Challenges in FER: Variety of masks depicted in the right-hand columns of (a) and (b), and (a)stonger expressions are easier for FER while (b)subtler ones are hard. }
  \label{fig:EvD}
\end{figure}
While FER in the presence of masks is challenging enough, the issue of data adds to the difficulty;  Datasets for occlusion are few in number and small in size. Further, none are specific to the case of face masks. While researchers are working on handling occlusion in FER, little has been done in the past to develop occlusion (masked) datasets. Wang et al. \cite{Wang2020RegionAN} segregated occlusion images from three popular FER datasets: FERPlus \cite{inproceedings}, RAFDB \cite{li2017reliable, li2019reliable} and and AffectNet \cite{affect}, to form Occlusion-FERPlus, Occlusion-RAFDB, and Occlusion-AffectNet. FEDRO \cite{8576656} is another recently released FER dataset with real-world occlusion. The above datasets have occlusion by sunglasses, medical masks, hands, and hair. While the number of images in these datasets is small, mask occluded images are present only in FEDRO dataset. \autoref{tab:stats1} presents the statistics of the occlusion datasets mentioned. Towards masked FER, Yang et al. \cite{abc} manually crawled 562 masked FER images from the internet to obtain M-FER-T dataset comprising masked images. However, the dataset is annotated with three labels - positive, negative, and neutral - rather than the seven basic emotions used in popular datasets like RAFDB \cite{li2017reliable, li2019reliable} and AffectNet \cite{affect}. 

\begin{figure*}
  \centering
  \includegraphics[scale = 0.5]{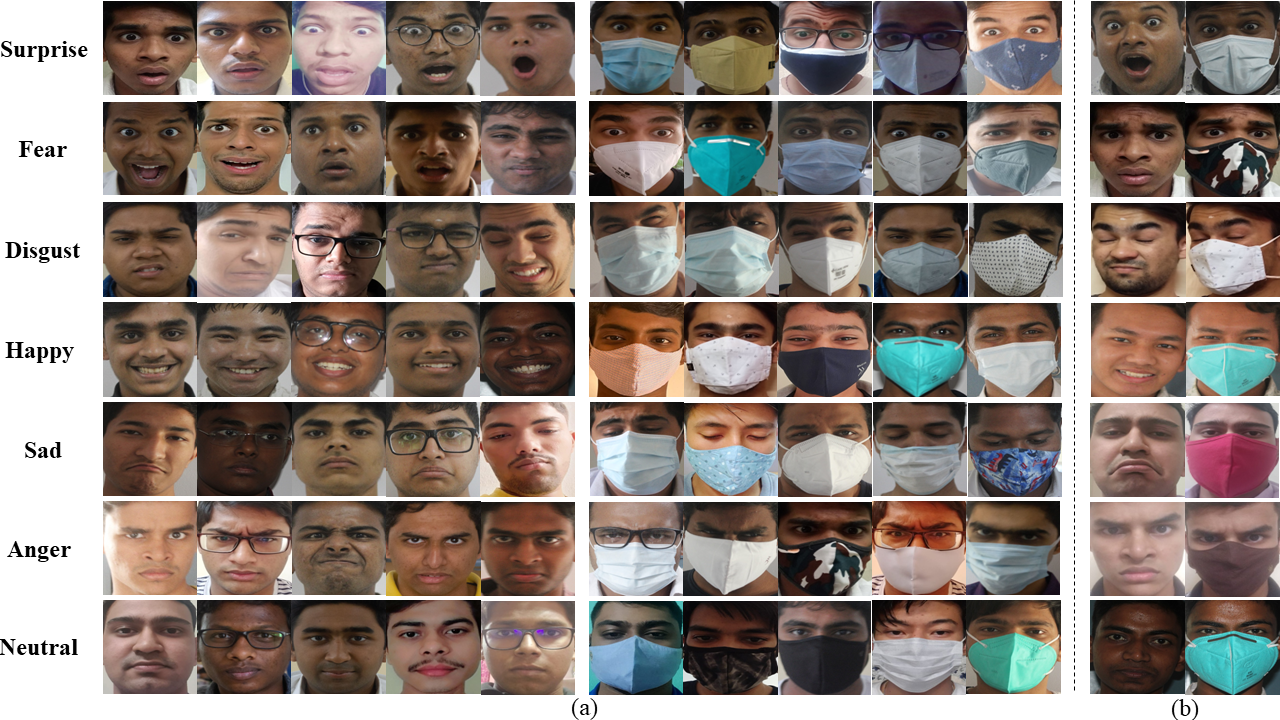}
  \caption{Sample images from (a)MSD-E and (b)MSD-PE. Non-masked images are on the left and masked are on the right.}
  \label{fig:examples}
\end{figure*}

Currently, work on masked FER is benchmarked on synthetically created datasets. In \cite{abc}, 10,487 images from LFW dataset are  annotated with three expression classes as before to get M-LFW-FER. They also created M-KDDI-FER -- a dataset with 17,236 images. Gera et al. \cite{articleG} synthetically applied face masks using a publicly available tool \footnote{https://github.com/X-zhangyang/Real-World-Masked-Face-Dataset} on FERPlus \cite{inproceedings} and RAFDB \cite{li2017reliable, li2019reliable} to obtain Masked-FERPlus, which has 28,709 images for training and 3589 images for testing, and Masked-RAFDB, which has 12,271 and 3,008 images for training and testing respectively. 

Apart from the above-mentioned real-world occlusion datasets and the synthetic masked datasets, we could not find any other study of masked FER datasets. This striking lack of data has motivated us to propose the \textbf{Masked Student Dataset of Expressions (MSD-E)}. The dataset consists of 1,960 images annotated with the 7 basic emotions: Surprise, fear, disgust, happy, sad, anger, and neutral. 142 subjects provided non-masked and masked images of these expressions. To ensure that the labels are reliable, we ask volunteers to verify whether the expression label given during image collection matches the expression performed. Based on the observations, we relabel any images with a disparity.

We then benchmark our dataset using ResNet-18 \cite{inproceedings} and two SOTA methods for occlusion in FER. To further tackle the case of occlusion by masks in FER, we employ two training paradigms on our dataset: Contrastive Learning (CL) and Knowledge Distillation (KD). These methods are motivated by work done in the field of masked face recognition \cite{NetoFocus21, GeorgeTeach21, HuberMask21}. As both paradigms require pairs of images -- one non-masked and one masked, we test them on a subset of the expression dataset, MSD-E, called Masked Student Dataset of Paired Expressions (MSD-PE) that consists of pairs of non-masked and masked images. Our experiments show that these training paradigms significantly increase FER accuracy in the masked case while dampening the drop in performance in the non-masked case. They also increase the model's discrimination power in the case of hard emotions like disgust and fear. Significantly, the dataset has numerous hard images, i.e., images where the expression is subtle or a mixture of emotions as depicted in \autoref{fig:EvD}.
\begin{table}[hbt!]
    \centering
    \caption{No. of Images in Challenging Subsets of Benchmark Datasets}
    \begin{tabular}{ccc}
    \toprule
       Dataset  &  Occluded Images & Masked Images\\
       \cmidrule(r){1-3}
        FERPlus \cite{inproceedings} & 111 & 0 \\
        RAFDB \cite{li2017reliable, li2019reliable} & 123 & 0 \\
        Affectnet \cite{affect} & 100 & 0 \\
        FEDRO \cite{8576656} & 234 & 10 \\
        \bottomrule
    \end{tabular}
    \label{tab:stats1}
\end{table}
To the best of our knowledge, the expression dataset MSD-E is the first FER dataset collected specifically for the masked scenario and the first FER dataset of Indian ethnicity. The following sections discuss the dataset in detail and provide the specifics of experiments using the dataset.

\section{Masked Student Dataset of Expressions}

\subsection{Image Data}

The basic facial expressions of 142 volunteers were recorded using two devices. The first was a DSLR camera with 31.7 megapixels and a $36 \times 24$ mm sensor, while the second was a mobile phone camera with a 50 megapixels (f/1.8) primary camera and two 2 megapixels (f/2.4) cameras. We do not experiment with images collected from different cameras separately. The experimenter instructed each participant to perform the seven basic emotions: Surprise, fear, disgust, happy, sad, anger, and neutral, first without a mask and then with one. As a result, we obtained 14 images from each participant -- 7 non-masked and 7 masked. We anticipated that some images would have to be either removed due to camera artifact or relabeled while verifying the given labels. To avoid loss of data, we asked 14 participants with the ability to emote the expressions well to provide an additional set of images. The experimenter intentionally requested them to perform "tough" expressions such as fear, disgust, and anger. As a result, we obtained 1,007 pairs of images or 2,014 images in total before the verification process. 

The participants were all male, except for two, and their ages varied between 10 and 40. All the participants were of Indian ethnicity. Significantly, we had participants from all six physiographic divisions of India. In addition, the participant used their own mask. Some were the common surgical mask, while many used masks with printed designs, providing real variety in the type of mask in the images. Further, there is significant intra-class variation as different participants emoted the same expression differently. We discuss the challenges in the dataset in Section \ref{cal}. Since the dataset comprises primarily of student subjects, we name the dataset \textbf{M}asked \textbf{S}tudent \textbf{D}ataset of \textbf{E}xpessions or \textbf{MSD-E}.
\begin{table}[hbt!]

  \caption{Statistics of MSD-E and its subset MSD-PE.}
  \label{tab:freq}
  \begin{tabular}{ccccc}
    \toprule
    \multirow{2}{*}{\textbf{Expression}}
        &\multicolumn{3}{c}{\textbf{MSD-E}}  
        
    &
    \multirow{2}{*}{\textbf{MSD-PE}(Pairs)} \\
    \cmidrule(r){2-4}

         & Non-masked     & Masked    & Total \\
         \hline
    
    Surprise        & 128                & 121       &249 &113\\
    Fear        & 121 & 128 & 249      &108\\
    Disgust     & 114 & 127 & 241         &106\\
    Happy & 193 & 164 & 357 &155\\
    Sad & 123 & 142 & 256 &119\\
    Angry & 104 & 143 & 247 &100\\
    Neutral & 181 & 171 & 352 &156\\
    \hline
    \textbf{Total} & \textbf{964} & \textbf{996} & \textbf{1960} &\textbf{857}\\
\bottomrule
\end{tabular}
\end{table}

\subsection{Validating Expression Labels} \label{val}

A major limitation of posed facial expression data is that the expression requested may not always match the expression performed. This is because the subject may be unable to act out an otherwise spontaneous facial expression. Assuming these expression labels to be the ground truth is highly unreliable. Taking a leaf from the work of Lucey et al. \cite{5543262} we determined whether (1) the image matched the given label, (2) whether the image needed to be relabelled or, (3) whether it needed to be removed. The verification process was as follows:
\begin{enumerate}
\item 7 volunteers used perceptual judgment to determine whether an image matched its expected expression class. Working independently, they visually inspected the images and flagged any image they judged did not lie in its given class. If possible, they marked another class the image could belong to, giving it a new label. If an image was not flagged by the volunteer, its new label was its original label. 
\item After collating the volunteers' remarks, majority voting based on the new label determined the action performed on the image. Images not flagged by anyone were trivially reassigned to their original label. 1,719 images were not flagged by any participant. For the remaining 288 images flagged by some number of participants, we noted the new label with the highest number of endorsements. The image was assigned that class if at least half the participants approved of that label and no more than 25 percent endorsed another class. The images with their new labels made it to the \textbf{Mask Student Dataset of Expressions (MSD-E)}. 
\item To obtain the \textbf{Mask Student Dataset of Paired Expressions (MSD-PE)}, we had to ensure that the non-masked image and its corresponding masked image had the same new label. To this end, we removed all pairs of images that had a mismatch. 150 pairs were removed, leaving us 857 pairs of non-masked and masked images.
\end{enumerate}
The inventory of the above validation process for the two datasets is given in \autoref{tab:freq} and examples of image pairs are depicted in \autoref{fig:examples}.

\subsection{Challenges in the Dataset} \label{cal}
FER in the real world will need to be nuanced. Machines will need to be robust to the presence of eyewear, like spectacles and sunglasses. For this reason, we did not ask subjects to remove their glasses during data collection. Apart from this, the variety of masks and their styles, and the manner in which expressions are emoted contribute to the complexity of real-world FER. We discuss how our dataset incorporates these challenges below.

\noindent\textbf{Variety of masks: }The masked example images in \autoref{fig:examples} depict the variety of masks used by the volunteers while posing for the photos. A sizeable number used the common light blue surgical mask; another substantial set of participants brought with them an N95 mask that was either white, blue, or black; many more participants used a cloth mask. The cloth masks contribute the most variety to the dataset as they come in different colours and designs. Our dataset reflects the real-world scenario concerning masks used in images. But this is not the only way our dataset exhibits real-world attributes. 

\noindent\textbf{Intra-class variation: }As discussed earlier, facial expressions reflect our thoughts and feelings. There cannot be discrete levels or combinations of states of being. Consequently, the representations of these states through facial expressions cannot be disconnected from one another; they must be combinational. For precisely this reason, datasets like RAFDB \cite{li2017reliable, li2019reliable} employ compound emotion labelling where each image is labeled with multiple emotions to represent the expression better. For MSD-E, we have chosen to label each image with the dominant expression. However, this does not take away from the fact that many images have more than one emotion present. For instance, some images may be best described as "fearfully surprised" or "angrily disgusted" and would have been labeled as surprise and disgust, respectively. To add to this, it may be easy to classify some images compared to others. Images where the dominant expression is "harsh" and "loud" are easy for a human and a machine to distinguish. On the other hand, the expression in other images is subtler and may be confusing for even humans to discern. However, such expressions arise from the sundry of facial expressions, making it imperative that computer vision models learn to perform on easy and difficult expression images to enable a more nuanced human-computer interaction. \autoref{fig:EvD} depicts examples of easy and difficult images in our dataset. 

With such variety, our dataset presents various challenges. But these challenges provide opportunities to fine-tune FER models to perform well in a real-world environment. Along with our extensive validation process mentioned in Section \ref{val}, we present a novel dataset with real-world subtleties and characteristics. The following section discusses experiments we have performed using MSD-E.

\section{Experiments}

For robust FER in the current post-pandemic scenario, we need models that perform well in the non-masked and masked case simultaneously. While FER architectures have been developed to be occlusion-invariant \cite{articleG,Wang2020RegionAN,8576656,ding2020occlusion}, their performance has not been tested on real-world masked expression data. For these reasons, we perform the following experiments:
\begin{itemize}
    \item We find the performance of ResNet-18 \cite{inproceed} on MSD-E and MSD-PE to act as a baseline to compare to other models.
    \item  We evaluate the performance of two training paradigms: Contrastive learning \cite{NetoFocus21} and  knowledge distillation \cite{GeorgeTeach21, HuberMask21}, that have both shown promising results in masked face recognition. As these paradigms require pairs of images, we evaluate their performance on MSD-PE.
    \item Finally, we evaluate the performances of state-of-the-art (SOTA) methods on MSD-E and MSD-PE, and compare them with that of contrastive learning and knowledge distillation. These are particularly approaches that have been developed to tackle the problem of occlusion in FER.
\end{itemize}

\subsection{Experimental Setup} 

\textbf{Training Protocol: }
While performing experiments on MSD-E or MSD-PE, we use 10-fold cross validation strategy to make use of all the data. To this end, the images are separated into 10 identity-independent folds. One fold is set as the validation set while the remaining are used for training. The model's accuracy on the validation set is noted. This is done 10 times, once with each fold as the validation set. We then report the average validation set accuracy. As the number of images per subject is not the same, the number of images per fold varies slightly as well.

\noindent\textbf{Implementation Details: }
ResNet-18 is used as the backbone network for all experiments, except when mentioned otherwise. Pre-trained weights of ResNet-18 trained on large scale face recognition dataset MS-Celeb-1M \cite{msceleb} are loaded onto the backbone. We use Adam optimizer and Cross Entropy loss for classification. The models are trained for 25 epochs with an initial learning rate of 0.0001 and 0.001 for the backbone and classification layer respectively which is reduced by a factor of 10 at epochs 8, 14, and 20. All the experiments are implemented in PyTorch and are run on a Tesla P100 GPU. 
\begin{figure*}
  \centering
  \includegraphics[scale = 0.5]{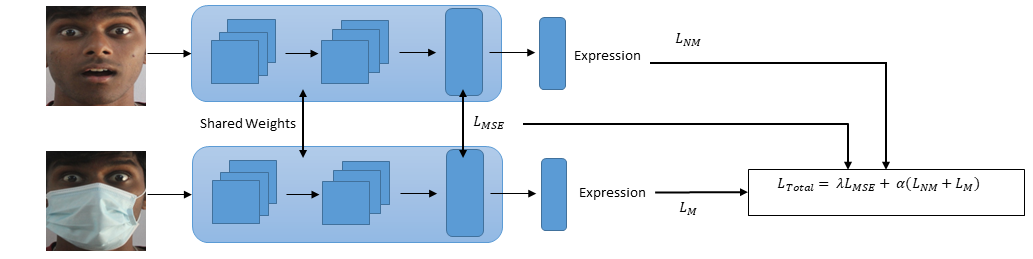}
  \caption{Architecture of Contrastive Learning-based FER on MSD-PE. Two feature vectors are generated by passing a non-masked and masked image to the model. Their difference is given by \textbf{$L_{\mathrm{MSE}}$}. The error in expression classification is given by \textbf{$L_{\mathrm{NM}}$} and \textbf{$L_{\mathrm{M}}$}. The architecture is based on the work of Neto et al.[15].} 
  \label{fig:CL}
\end{figure*}
\subsection{Baseline Results} \label{base}
In order to establish the baseline performance on MSD-E, we trained and tested ResNet-18 \cite{inproceed} on three subsets of MSD-E using the protocol mentioned in the previous subsection. The first was the Non-masked MSD-E (represented as \textbf{NM}), comprising of only non-masked images; the second was Masked MSD-E (represented as \textbf{M}), comprising of only masked images; the third was the whole MSD-E (represented as \textbf{Mix}). The classification accuracies obtained are reported in \autoref{tab:accs1} for NM and \autoref{tab:accs12} for M. The model acheives 65.35\% accuracy when trained and tested on NM, and 45.85 \% when trained and tested on M using the training protocol mentioned. However, when trained on Mix, the model's performance on both non-masked \textbf{(55.88\%)} and masked images \textbf{(44.15\%)} dips significantly.  These results indicate that the model is unable to learn FER for both non-masked and masked images concurrently when trained routinuely on the whole set of images as it "forgets" FER for the non-masked case. To avoid this "forgetting", we require paradigms that assist the model to improve its masked accuracy while retaining its non-masked performance. The next subsection discusses two approaches that intend to accomplish this. For comparison of these methods, we compute the baseline on MSD-PE which is reported in \autoref{tab:accs2}.

\subsection{Experiments on MSD-PE}
As mentioned earlier, we need mask-invariant models that perform well on non-masked images as well. The results in Section \ref{base} tell us that routinue training does not achieve this. Two methods that have shown promising results in masked face recognition literature are Contrastive Learning (CL) \cite{NetoFocus21} and Knowledge Distillation (KD) \cite{GeorgeTeach21, HuberMask21}. We evaluate their performance for FER using the MSD-PE dataset. We note here that we use the whole MSD-PE dataset, i.e., both non-masked and masked images at the same time. First, we briefly describe the approaches. Both CL and KD aim to promote the model to produce similar feature vectors for both non-masked and masked images. The intuition is that this will force the model to focus on non-occluded regions of the face. However, CL and KD do this in different ways. We discuss the details below. As far as we are aware, this is the first study on the use of CL and KD for FER in the face mask scenario.\newline

\noindent\textbf{Contrastive Learning: }
In the case of contrastive learning, two images are passed to the model: one is a non-masked image of subject A with expression E, the other is the corresponding masked image, i.e., the image of subject A expressing E with a mask on. The model produces two feature embeddings of size $7\times7\times512$, one for each image. The model is trained to produce non-masked and masked feature vectors that are closer to each other using Mean Square Error loss given below. \begin{equation} \label{eq1}
\mathbf{L_{\mathrm{MSE}}} = \frac{1}{B}\frac{1}{F}\Sigma_{i=1}^{B}{\Sigma_{j=1}^{F}{(xn_{\mathrm{ij}} - xm_{\mathrm{ij}})^2}}
\end{equation}
Here, B represents the mini-batch size, F is the feature size, $xn_{\mathrm{ij}}$  is the value of the non-masked feature vector i at position j, while  $xm_{\mathrm{ij}}$ represents the same but for the masked image feature vector. To train the network for FER, standard cross entropy loss is used. In Eq. 2, M stands for the number of emotion classes, $y_c$ is the binary indicator 1 if class label is class c and 0 otherwise, while $p_{o,c}$ is the predicted probability that observation o belongs to class c. 
\begin{equation} \label{eq2}
\mathbf{L_{\mathrm{CE}}} = -\sum_{c=1}^My_{o,c}\log(p_{o,c})
\end{equation}
We use two $L_{\mathrm{CE}}$ losses: one for non-masked images, and another for masked images denoted as $L_{\mathrm{NM}}$ and $L_{\mathrm{M}}$ respectively. The total loss is combination of the three losses mentioned above as given in Eq. 3. Here, $\lambda$ and $\alpha$ control how much each loss contributes to training and we set them to 0.2 and 1.0 respectively. The architecture is depicted in \autoref{fig:CL}. 
\begin{equation} \label{eq3}
\mathbf{L_{\mathrm{Total}}^\text{CL}} = \lambda\mathbf{L_{\mathrm{MSE}}} + \alpha(L_{\mathrm{NM}} + L_{\mathrm{M}})
\end{equation}
Using the experimental setup and training protocol mentioned before, we train ResNet-18 \cite{inproceed}, employing CL, on MSD-PE and report the non-masked and masked results obtained. These results are presented in \autoref{tab:accs2} and the mean diagonal values of the confusion plots are available in \autoref{tab:diag}. From these results, we find that CL achieves its primary objective of bettering masked accuracy with an increase of 3.09\% compared to the baseline. In addition, non-masked accuracy increased by 4.13\% compared to the baseline. It is worth noting that this has been achieved with no increase in the number of trainable parameters. \newline
 
  \begin{table}[hbt!]
\caption{FER accuracy (\%)  on NM (non-masked images of MSD-E)}
  \centering
  \begin{tabular}{cc|c}
    \toprule
    Method    & Non-Masked  & No. of parameters\\
    \midrule
    Baseline on NM  &65.35 & $\sim$11M\\

    RAN \cite{Wang2020RegionAN} &61.55 &$\sim$23M  \\
    SCAN \cite{articleG} &66.41 &$\sim$66M  \\
    \bottomrule
  \end{tabular}
  \label{tab:accs1}

\end{table}

  \begin{table}[hbt!]
\caption{FER accuracy (\%)  on M (masked images of MSD-E)}
  \centering
  \begin{tabular}{cc}
    \toprule
    Method    & Accuracy \\
    \midrule
    Baseline &45.854 \\

    RAN \cite{Wang2020RegionAN} &39.41  \\

    SCAN \cite{articleG} &49.29  \\
    \bottomrule
  \end{tabular}
  \label{tab:accs12}

\end{table}

\begin{table}[hbt!]
\caption{FER accuracy (\%)  on MSD-PE. Models are trained on whole of MSD-PE. Results are given for different subsets.}
  \centering
  \begin{tabular}{ccc|c}
    \toprule
    Method    & Non-Masked & Masked  & Average\\
    \midrule
    Baseline  &56.25 &43.71 &49.98 \\
    CL & 60.01 & 48.06 & 54.03\\
    KD & 57.63 & 45.86  & 51.74\\
    RAN \cite{Wang2020RegionAN} &60.89 &41.54  &51.42   \\
    SCAN \cite{articleG} &65.01 &57.29 &61.15  \\
    \bottomrule
  \end{tabular}
  \label{tab:accs2}

\end{table}

 \begin{table*}[hbt!]

 \caption{Mean diagonal value of confusion matrices.}
  \centering
  \begin{tabular}{llllllll}
    \toprule
    \multirow{2}{*}{\textbf{Method}} & 
    \multicolumn{5}{c}{\textbf{Expression}}                   \\
    \cmidrule(r){2-8}
      &Surprise& Fear & Disgust & Happy & Sad & Angry & Neutral \\
      \hline
      Train on NM  &0.78 &0.46 &0.57 &0.81 &0.50 &0.59 &0.95 \\
      Train on M   &0.73 &0.15 & 0.32 &0.47 &0.34 & 0.43 & 0.56 \\
      Train on Mix &0.72 & 0.27 &0.39 &0.68 &0.45 &0.48 &0.66 \\
      \midrule
      CL           &0.61 &0.17 &0.55 &0.56 &0.3 &0.43 &0.77 \\
      KD           &0.67 &0.16 &0.52 &0.53 &0.48 &0.38 &0.62 \\

    \bottomrule
  \end{tabular}
  \label{tab:diag}

\end{table*}

\begin{figure*}
  \centering
  \includegraphics[scale = 0.5]{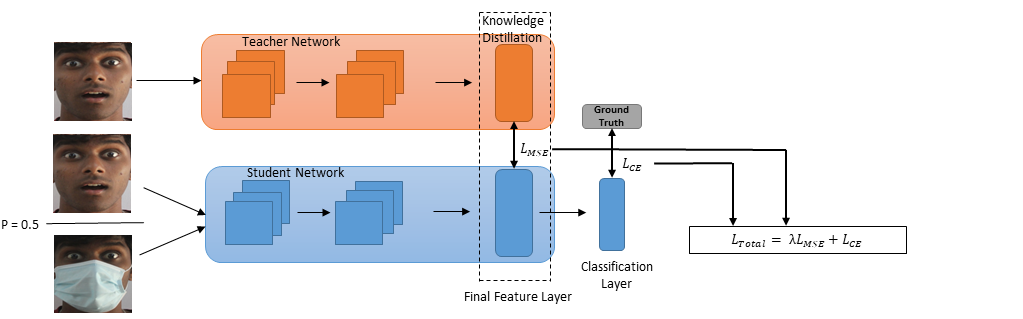}
  \caption{Architecture of Knowledge Distillation-based FER on MSD-PE. A non-masked or masked image is randomly passed to the model. The corresponding non-masked image is passed to the teacher model. \textbf{$L_{\mathrm{MSE}}$} gives the difference between the feature vectors obtained. \textbf{$L_{\mathrm{CE}}$} gives the classification error of the student model. The student model is trained using \textbf{$L_{\mathrm{Total}}$}.}
  \label{fig:KD}
\end{figure*}

\textbf{Knowledge Distillation: }
While their objectives are similar, knowledge distillation differs from contrastive learning as it makes use of a teacher network that is already trained on non-masked images for FER. A student network "learns" masked FER from the teacher network by moving the feature vectors of the student model closer to those of the teacher model. During training, a non-masked image is fed to the teacher model. On the other hand, the student model is randomly given, with a probability threshold 'p',  either the same non-masked image, or its corresponding masked image. This is done to ensure that the student model performs well on both non-masked and masked images. We set 'p' as 0.5. The feature vectors produced by the student model and the teacher model are pushed closer using Mean Square Error given in Eq 1. The student model is trained for FER using the Cross Entropy Loss given in Eq 2. The total loss is formally defined in Eq 4 where $\lambda$ controls the amount of knowledge distillation from the teacher model. We set $\lambda$ to 100.  A diagrammatic overview of the proposed training framework is presented in \autoref{fig:KD}. 
\begin{equation} \label{eq4}
\mathbf{L_{\mathrm{Total}}^\text{KD}} = \lambda\mathbf{L_{\mathrm{MSE}}} + L_{\mathrm{CE}}
\end{equation}

The result of using KD for FER are presented in \autoref{tab:accs2} and the mean diagonal values of confusion plots, indicating the specific discrimination ability of the model for each expression, are given in \autoref{tab:diag}. KD performs above the baseline but does not perform better than CL. Its accuracy is 2.15\% better than the baseline in the masked scenario and 1.38\% better in the non-masked case. The number of trainable parameters in KD is still the same: 11M. The only addition are the frozen weights of the teacher model, meaning there are still 11M parameters during inference.

\subsection{Visualizations}
To further evaluate the impact of the techniques used above, we use two types of visualizations. 

\begin{figure*}
  \centering
  \includegraphics[scale = 0.5]{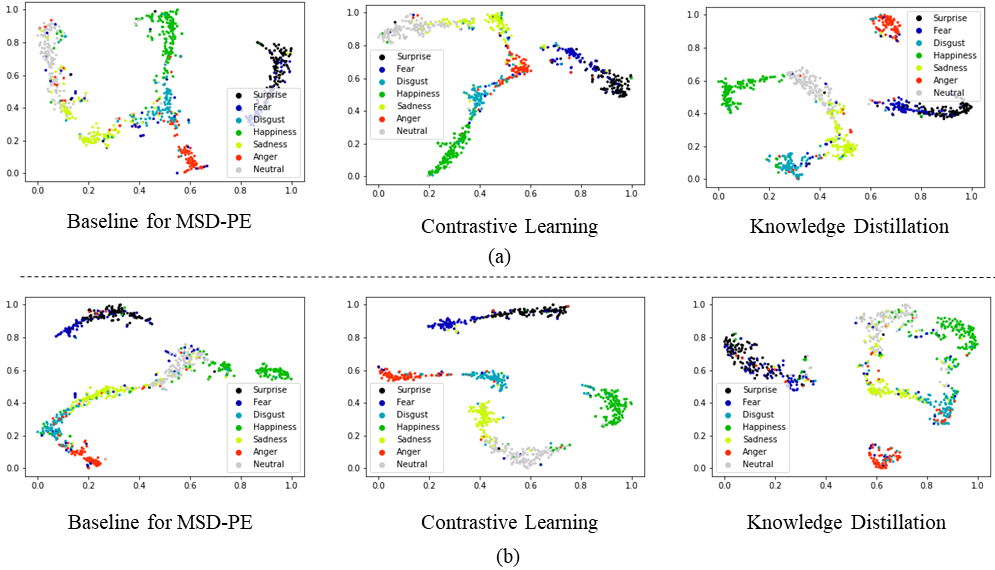}
  \caption{t-SNE plots of (a) Non-masked features and (b) Masked features for different experiments. Best viewed in colour.}
  \label{fig:tsne}
\end{figure*}

\textbf{t-SNE plots: }We first use t-SNE plots \cite{tsne} to visualise the distribution of the non-masked and masked facial expression features . As seen in \autoref{fig:tsne}(a), the non-masked features produced by the baseline are not well-separated. In particular, the features of the anger and fear classes are mixed with other classes. On the other hand, the features produced by CL and KD are more sparse. Coming to masked features, CL provides the most discriminative facial expression features, with evident intra-class similarity as seen in \autoref{fig:tsne}(b). KD, however, does not give better-separated features compared to the baseline in this case.

\begin{figure*}
  \centering
  \includegraphics[scale = 0.5]{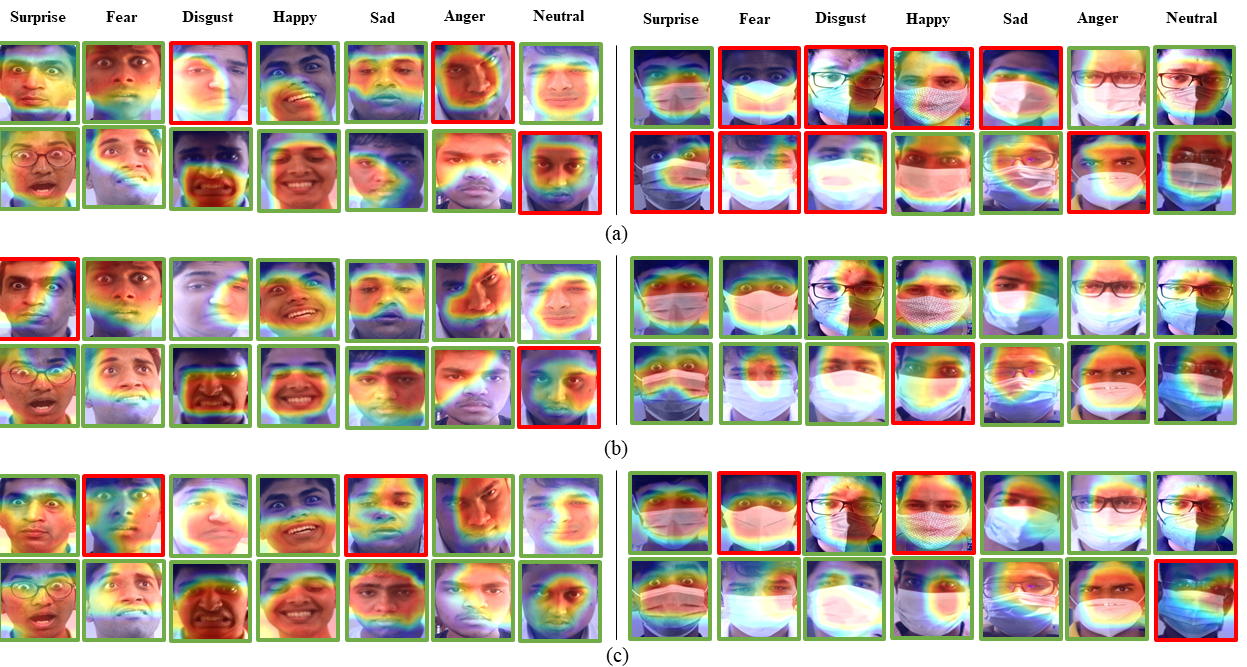}
  \caption{GradCam Visualization on (a) Baseline for MSD-PE, and using (b) Contrastive Learning and (c) Knowledge Distillation. Non-masked images are on the left and masked images on the right. Green and red boxes around images indicate whether the model got the prediction right or wrong respectively.The heatmap is superimposed over the image, where areas highlighted in red contributed the most toward classification. Best viewed in colour.}
  \label{fig:fig1}
\end{figure*}

\textbf{Grad-CAM: }To test our hypothesis that CL and KD push the model to focus on non-occluded regions of the face, we employ Grad-CAM\cite{8237336}. \autoref{fig:fig1} depicts the result of applying Grad-CAM to non-masked and masked images. In the masked case, we see that when the model is trained on both types of images routinely, it focuses on occluded areas at times. As seen in \autoref{fig:fig1}(a), a majority of images have a considerable portion of the mask highlighted, indicating it was used for FER. In the case of CL and KD (refer \autoref{fig:fig1}(b) and (c)), however, we find that the model prioritizes the eye and forehead region for decision making, and the focus on the mask is comparitively less.  It follows that CL and KD make better use of the limited features available in the masked FER scenario, justifying our choice to use them. In the non-masked scenario, we find that the models use relevant features from the entire face; the mouth region seems to be an important feature for detecting happy, surprise, disgust and neutral expressions, establishing the premise for this paper: masks occlude vital facial features for FER.  
\begin{figure*}
  \centering
  \includegraphics[width = \linewidth]{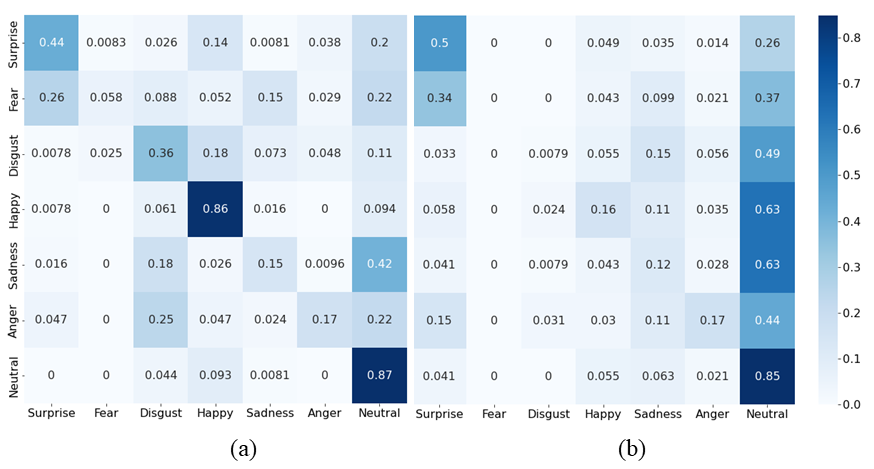}
  \caption{Confusion matrices of cross-dataset study. (a)RAFDB \cite{li2017reliable, li2019reliable} to non-masked images in MSD-E. (b)Synthetically masked RAFDB to masked images of MSD-E. Predicted labels are on the vertical axis and true labels are on the horizontal axis. }
  \label{fig:conf}
\end{figure*}
\subsection{Performance of state-of-the-art methods}
In this section, we report the performance of occlusion-robust SOTA FER methods, RAN \cite{Wang2020RegionAN} and SCAN \cite{articleG}, on MSD-E and MSD-PE. Region attention network (RAN) selects image crops from the input image independent of landmark points. A relation-attention module relates local and global context and gives the expression representation for classification. Our experiments with RAN \cite{Wang2020RegionAN} use ResNet-50 \cite{inproceed} as the backbone with fixed position cropping of images, resulting in six crops per image. Spatio-channel Attention Net Branch (SCAN) \cite{articleG} computes attention weight for each channel and spatial location within the channel across all local patches to make the model robust to occlusion. We use the publically available implementation of SCAN\footnote{https://github.com/1980x/SCAN-CCI-FER} for the experiments. We note here that we used weights pretrained on AffectNet for the backbone of SCAN. As before, we use 10-fold cross validation while evaluating these methods as well.  

\textbf{MSD-E}. \autoref{tab:accs1} and \autoref{tab:accs12} presents the FER accuracy of RAN \cite{Wang2020RegionAN} and SCAN \cite{articleG} on MSD-E along with the number of trainable parameters for each method. SCAN achieves 3.44 \% above the baseline on M. It also surpasses the baseline on NM by 1.06\%. RAN \cite{Wang2020RegionAN}, however does not surpass the baseline in either case. 

\textbf{MSD-PE}. \autoref{tab:accs2} presents the comparison on MSD-PE. Here too SCAN massively outperforms the baseline and CL and KD as well. It is worth noting that SCAN achieves 55\% on masked images, i.e., 7.4\% higher than CL but has 6 times more parameters and is loaded with AffectNet pretrained weights. On the other hand, RAN produces better results on non-masked images than CL and KD while falling short on masked images.

\subsection{Cross-Dataset Study}
To explore the universality of the expressions in our dataset, we conducted a RAFDB to MSD-E cross-dataset study. We used images of RAFDB for training and those of MSD-E as testing. In particular, we trained ResNet-18 on RAFDB and then tested it on the non-masked subset of MSD-E. We then, trained ResNet-18 on synthetically masked RAFDB images and tested them on the masked images in MSD-E. \autoref{fig:conf} depicts the confusion matrices of these two experiments. Surprise, disgust, happiness, and neutral were easier to recognise in the non-masked case, while fear was confused with surprise, as shown in \autoref{fig:conf}(a). In the masked case, depicted in \autoref{fig:conf}(b), however, only surprise and neutral were recognised easily and fear was confused with surprise again. This indicates that training on synthetically masked images for FER is not reliable, further emphasising the need for real-world masked FER data.
\section{Conclusion}
As society emerges from the Covid-19 pandemic, we have grown accustomed to wearing masks in public. FER systems, however, find expression recognition in this new normal an arduous task. This challenge is increased by the variety of masks people use and the high intra-class variation. In this paper, we have proposed a novel FER dataset pertaining to the face mask scenario called MSD-E with a subset consisting of pairs of images called MSD-PE. MSD-E contains 1,960 images and MSD-PE has 857 pairs of images. Each image is labeled for different facial expressions. We used contrastive learning and knowledge distillation to develop FER models that are mask-invariant and demonstrate how they make the model focus on visible areas in the face mask scenario. Using these training paradigms on larger networks can perhaps learn more discriminative features for FER. Finally, we tested SOTA methods on the dataset and performed a cross-dataset study. By releasing this dataset for research, we hope that more researchers will be encouraged to address the hard problem of drastic occlusion by masks in FER. Our aspiration is that the dataset will be a useful benchmark resource to compare FER solutions in this challenging condition.


\bibliographystyle{ACM-Reference-Format.bst}
\bibliography{references.bib}


\begin{thebibliography}{19}


\ifx \showCODEN    \undefined \def \showCODEN     #1{\unskip}     \fi
\ifx \showDOI      \undefined \def \showDOI       #1{#1}\fi
\ifx \showISBNx    \undefined \def \showISBNx     #1{\unskip}     \fi
\ifx \showISBNxiii \undefined \def \showISBNxiii  #1{\unskip}     \fi
\ifx \showISSN     \undefined \def \showISSN      #1{\unskip}     \fi
\ifx \showLCCN     \undefined \def \showLCCN      #1{\unskip}     \fi
\ifx \shownote     \undefined \def \shownote      #1{#1}          \fi
\ifx \showarticletitle \undefined \def \showarticletitle #1{#1}   \fi
\ifx \showURL      \undefined \def \showURL       {\relax}        \fi
\providecommand\bibfield[2]{#2}
\providecommand\bibinfo[2]{#2}
\providecommand\natexlab[1]{#1}
\providecommand\showeprint[2][]{arXiv:#2}

\bibitem[Barsoum et~al\mbox{.}(2016)]%
        {inproceedings}
\bibfield{author}{\bibinfo{person}{Emad Barsoum}, \bibinfo{person}{Cha Zhang},
  \bibinfo{person}{Cristian Ferrer}, {and} \bibinfo{person}{Zhengyou Zhang}.}
  \bibinfo{year}{2016}\natexlab{}.
\newblock \showarticletitle{Training Deep Networks for Facial Expression
  Recognition with Crowd-Sourced Label Distribution}.
  \bibinfo{pages}{279--283}.
\newblock
\urldef\tempurl%
\url{https://doi.org/10.1145/2993148.2993165}
\showDOI{\tempurl}


\bibitem[Bo et~al\mbox{.}(2020)]%
        {abc}
\bibfield{author}{\bibinfo{person}{Yang Bo}, \bibinfo{person}{Jianming Wu},
  {and} \bibinfo{person}{Gen Hattori}.} \bibinfo{year}{2020}\natexlab{}.
\newblock \bibinfo{title}{Facial Expression Recognition with the advent of Face
  Masks}.
\newblock
\newblock
\urldef\tempurl%
\url{https://doi.org/10.1145/3428361.3432075}
\showDOI{\tempurl}


\bibitem[Ding et~al\mbox{.}(2020)]%
        {ding2020occlusion}
\bibfield{author}{\bibinfo{person}{Hui Ding}, \bibinfo{person}{Peng Zhou},
  {and} \bibinfo{person}{Rama Chellappa}.} \bibinfo{year}{2020}\natexlab{}.
\newblock \showarticletitle{Occlusion-adaptive deep network for robust facial
  expression recognition}. In \bibinfo{booktitle}{\emph{2020 IEEE International
  Joint Conference on Biometrics (IJCB)}}. IEEE, \bibinfo{pages}{1--9}.
\newblock


\bibitem[Georgescu et~al\mbox{.}(2022)]%
        {GeorgeTeach21}
\bibfield{author}{\bibinfo{person}{Mariana-Iuliana Georgescu},
  \bibinfo{person}{Georgian-Emilian Du{\c{t}}ǎ}, {and}
  \bibinfo{person}{Radu~Tudor Ionescu}.} \bibinfo{year}{2022}\natexlab{}.
\newblock \showarticletitle{Teacher--student training and triplet loss to
  reduce the effect of drastic face occlusion}.
\newblock \bibinfo{journal}{\emph{Machine Vision and Applications}}
  \bibinfo{volume}{33}, \bibinfo{number}{1} (\bibinfo{year}{2022}),
  \bibinfo{pages}{1--19}.
\newblock


\bibitem[Gera and S(2021)]%
        {articleG}
\bibfield{author}{\bibinfo{person}{Darshan Gera} {and}
  \bibinfo{person}{Balasubramanian S}.} \bibinfo{year}{2021}\natexlab{}.
\newblock \showarticletitle{Landmark Guidance Independent Spatio-Channel
  Attention and Complementary Context Information based Facial Expression
  Recognition}.
\newblock \bibinfo{journal}{\emph{Pattern Recognition Letters}}
  \bibinfo{volume}{145} (\bibinfo{date}{02} \bibinfo{year}{2021}).
\newblock
\urldef\tempurl%
\url{https://doi.org/10.1016/j.patrec.2021.01.029}
\showDOI{\tempurl}


\bibitem[Guo et~al\mbox{.}(2016)]%
        {msceleb}
\bibfield{author}{\bibinfo{person}{Yandong Guo}, \bibinfo{person}{Lei Zhang},
  \bibinfo{person}{Yuxiao Hu}, \bibinfo{person}{Xiaodong He}, {and}
  \bibinfo{person}{Jianfeng Gao}.} \bibinfo{year}{2016}\natexlab{}.
\newblock \showarticletitle{Ms-celeb-1m: A dataset and benchmark for
  large-scale face recognition}. In \bibinfo{booktitle}{\emph{European
  conference on computer vision}}. Springer, \bibinfo{pages}{87--102}.
\newblock


\bibitem[He et~al\mbox{.}(2016)]%
        {inproceed}
\bibfield{author}{\bibinfo{person}{Kaiming He}, \bibinfo{person}{Xiangyu
  Zhang}, \bibinfo{person}{Shaoqing Ren}, {and} \bibinfo{person}{Jian Sun}.}
  \bibinfo{year}{2016}\natexlab{}.
\newblock \showarticletitle{Deep Residual Learning for Image Recognition}.
  \bibinfo{pages}{770--778}.
\newblock
\urldef\tempurl%
\url{https://doi.org/10.1109/CVPR.2016.90}
\showDOI{\tempurl}


\bibitem[Huber et~al\mbox{.}(2021)]%
        {HuberMask21}
\bibfield{author}{\bibinfo{person}{Marco Huber}, \bibinfo{person}{Fadi
  Boutros}, \bibinfo{person}{Florian Kirchbuchner}, {and}
  \bibinfo{person}{Naser Damer}.} \bibinfo{year}{2021}\natexlab{}.
\newblock \showarticletitle{Mask-invariant face recognition through
  template-level knowledge distillation}. In \bibinfo{booktitle}{\emph{FG
  2021}}. IEEE, \bibinfo{pages}{1--8}.
\newblock


\bibitem[Kotsia et~al\mbox{.}(2008)]%
        {KOTSIA20081052}
\bibfield{author}{\bibinfo{person}{Irene Kotsia}, \bibinfo{person}{Ioan Buciu},
  {and} \bibinfo{person}{Ioannis Pitas}.} \bibinfo{year}{2008}\natexlab{}.
\newblock \showarticletitle{An analysis of facial expression recognition under
  partial facial image occlusion}.
\newblock \bibinfo{journal}{\emph{Image and Vision Computing}}
  \bibinfo{volume}{26}, \bibinfo{number}{7} (\bibinfo{year}{2008}),
  \bibinfo{pages}{1052--1067}.
\newblock
\showISSN{0262-8856}
\urldef\tempurl%
\url{https://doi.org/10.1016/j.imavis.2007.11.004}
\showDOI{\tempurl}


\bibitem[Li and Deng(2019)]%
        {li2019reliable}
\bibfield{author}{\bibinfo{person}{Shan Li} {and} \bibinfo{person}{Weihong
  Deng}.} \bibinfo{year}{2019}\natexlab{}.
\newblock \showarticletitle{Reliable Crowdsourcing and Deep Locality-Preserving
  Learning for Unconstrained Facial Expression Recognition}.
\newblock \bibinfo{journal}{\emph{IEEE Transactions on Image Processing}}
  \bibinfo{volume}{28}, \bibinfo{number}{1} (\bibinfo{year}{2019}),
  \bibinfo{pages}{356--370}.
\newblock


\bibitem[Li et~al\mbox{.}(2017)]%
        {li2017reliable}
\bibfield{author}{\bibinfo{person}{Shan Li}, \bibinfo{person}{Weihong Deng},
  {and} \bibinfo{person}{JunPing Du}.} \bibinfo{year}{2017}\natexlab{}.
\newblock \showarticletitle{Reliable Crowdsourcing and Deep Locality-Preserving
  Learning for Expression Recognition in the Wild}. In
  \bibinfo{booktitle}{\emph{2017 IEEE Conference on Computer Vision and Pattern
  Recognition (CVPR)}}. IEEE, \bibinfo{pages}{2584--2593}.
\newblock


\bibitem[Li et~al\mbox{.}(2019)]%
        {8576656}
\bibfield{author}{\bibinfo{person}{Yong Li}, \bibinfo{person}{Jiabei Zeng},
  \bibinfo{person}{Shiguang Shan}, {and} \bibinfo{person}{Xilin Chen}.}
  \bibinfo{year}{2019}\natexlab{}.
\newblock \showarticletitle{Occlusion Aware Facial Expression Recognition Using
  CNN With Attention Mechanism}.
\newblock \bibinfo{journal}{\emph{IEEE Transactions on Image Processing}}
  \bibinfo{volume}{28}, \bibinfo{number}{5} (\bibinfo{year}{2019}),
  \bibinfo{pages}{2439--2450}.
\newblock
\urldef\tempurl%
\url{https://doi.org/10.1109/TIP.2018.2886767}
\showDOI{\tempurl}


\bibitem[Lucey et~al\mbox{.}(2010)]%
        {5543262}
\bibfield{author}{\bibinfo{person}{Patrick Lucey}, \bibinfo{person}{Jeffrey~F.
  Cohn}, \bibinfo{person}{Takeo Kanade}, \bibinfo{person}{Jason Saragih},
  \bibinfo{person}{Zara Ambadar}, {and} \bibinfo{person}{Iain Matthews}.}
  \bibinfo{year}{2010}\natexlab{}.
\newblock \showarticletitle{The Extended Cohn-Kanade Dataset (CK+): A complete
  dataset for action unit and emotion-specified expression}. In
  \bibinfo{booktitle}{\emph{2010 IEEE Computer Society Conference on Computer
  Vision and Pattern Recognition - Workshops}}. \bibinfo{pages}{94--101}.
\newblock
\urldef\tempurl%
\url{https://doi.org/10.1109/CVPRW.2010.5543262}
\showDOI{\tempurl}


\bibitem[Mollahosseini et~al\mbox{.}(2017)]%
        {affect}
\bibfield{author}{\bibinfo{person}{Ali Mollahosseini}, \bibinfo{person}{Behzad
  Hasani}, {and} \bibinfo{person}{Mohammad Mahoor}.}
  \bibinfo{year}{2017}\natexlab{}.
\newblock \showarticletitle{AffectNet: A New Database for Facial Expression,
  Valence, and Arousal Computation in the Wild}.
\newblock \bibinfo{journal}{\emph{IEEE Transactions on Affective Computing}}
  (\bibinfo{year}{2017}).
\newblock


\bibitem[Neto et~al\mbox{.}(2021)]%
        {NetoFocus21}
\bibfield{author}{\bibinfo{person}{Pedro~C Neto}, \bibinfo{person}{Fadi
  Boutros}, \bibinfo{person}{Jo{\~a}o~Ribeiro Pinto}, \bibinfo{person}{Naser
  Darner}, \bibinfo{person}{Ana~F Sequeira}, {and} \bibinfo{person}{Jaime~S
  Cardoso}.} \bibinfo{year}{2021}\natexlab{}.
\newblock \showarticletitle{FocusFace: Multi-task contrastive learning for
  masked face recognition}. In \bibinfo{booktitle}{\emph{2021 16th IEEE
  International Conference on Automatic Face and Gesture Recognition (FG
  2021)}}. IEEE, \bibinfo{pages}{01--08}.
\newblock


\bibitem[Saleem et~al\mbox{.}(2021)]%
        {article1}
\bibfield{author}{\bibinfo{person}{Sharmeen Saleem}, \bibinfo{person}{Subhi
  Zeebaree}, {and} \bibinfo{person}{Maiwan Abdulrazzaq}.}
  \bibinfo{year}{2021}\natexlab{}.
\newblock \showarticletitle{Real-life Dynamic Facial Expression Recognition: A
  Review}.
\newblock \bibinfo{journal}{\emph{Journal of Physics: Conference Series}}
  \bibinfo{volume}{1963} (\bibinfo{date}{07} \bibinfo{year}{2021}),
  \bibinfo{pages}{012010}.
\newblock
\urldef\tempurl%
\url{https://doi.org/10.1088/1742-6596/1963/1/012010}
\showDOI{\tempurl}


\bibitem[Selvaraju et~al\mbox{.}(2017)]%
        {8237336}
\bibfield{author}{\bibinfo{person}{Ramprasaath~R. Selvaraju},
  \bibinfo{person}{Michael Cogswell}, \bibinfo{person}{Abhishek Das},
  \bibinfo{person}{Ramakrishna Vedantam}, \bibinfo{person}{Devi Parikh}, {and}
  \bibinfo{person}{Dhruv Batra}.} \bibinfo{year}{2017}\natexlab{}.
\newblock \showarticletitle{Grad-CAM: Visual Explanations from Deep Networks
  via Gradient-Based Localization}. In \bibinfo{booktitle}{\emph{2017 IEEE
  International Conference on Computer Vision (ICCV)}}.
  \bibinfo{pages}{618--626}.
\newblock
\urldef\tempurl%
\url{https://doi.org/10.1109/ICCV.2017.74}
\showDOI{\tempurl}


\bibitem[Van~der Maaten and Hinton(2008)]%
        {tsne}
\bibfield{author}{\bibinfo{person}{Laurens Van~der Maaten} {and}
  \bibinfo{person}{Geoffrey Hinton}.} \bibinfo{year}{2008}\natexlab{}.
\newblock \showarticletitle{Visualizing data using t-SNE.}
\newblock \bibinfo{journal}{\emph{Journal of machine learning research}}
  \bibinfo{volume}{9}, \bibinfo{number}{11} (\bibinfo{year}{2008}).
\newblock


\bibitem[Wang et~al\mbox{.}(2020)]%
        {Wang2020RegionAN}
\bibfield{author}{\bibinfo{person}{K. Wang}, \bibinfo{person}{Xiaojiang Peng},
  \bibinfo{person}{Jianfei Yang}, \bibinfo{person}{Debin Meng}, {and}
  \bibinfo{person}{Yu Qiao}.} \bibinfo{year}{2020}\natexlab{}.
\newblock \showarticletitle{Region Attention Networks for Pose and Occlusion
  Robust Facial Expression Recognition}.
\newblock \bibinfo{journal}{\emph{IEEE Transactions on Image Processing}}
  \bibinfo{volume}{29} (\bibinfo{year}{2020}), \bibinfo{pages}{4057--4069}.
\newblock


\end{thebibliography}

\appendix

\end{document}